\documentclass[10pt, conference]{IEEEtran}

\usepackage{amsmath,amssymb,amsfonts}
\usepackage{graphicx}
\usepackage{pifont}

\usepackage{amsthm}
\usepackage{xcolor}
\usepackage[figuresright]{rotating}

\usepackage{graphicx}
\graphicspath{{/}{fig/}}

\usepackage{array}
\usepackage{textcomp}
\usepackage{xcolor}
\usepackage{multirow}
\usepackage{booktabs}

\usepackage{mathtools}
\usepackage{breqn}
\usepackage{float}

\usepackage{pgfplots}
\pgfplotsset{compat=1.7}
\usepgfplotslibrary{groupplots}

\usepackage{multirow,tabularx}
\usepackage{hyperref}
\usepackage{flushend}

\usepackage{setspace}

\newlength\figureheight
\newlength\figurewidth
\title{
    Applications of UWB Networks and Positioning to Autonomous Robots and Industrial Systems \\
}

\author{
    \IEEEauthorblockN{
        \vspace{1em}
        Yu Xianjia\IEEEauthorrefmark{3},
        Li Qingqing\IEEEauthorrefmark{3},
        Jorge Peña Queralta\IEEEauthorrefmark{3},
        Jukka Heikkonen\IEEEauthorrefmark{3},
        Tomi Westerlund\IEEEauthorrefmark{3}
    }
    \IEEEauthorblockA{
        \normalsize
        \IEEEauthorrefmark{3}\href{https://tiers.utu.fi}{Turku Intelligent Embedded and Robotic Systems (TIERS) Lab, University of Turku, Finland}.\\
        Emails: \textsuperscript{1}\{xianjia.yu, qingqli, jopequ, jukhei, tovewe\}@utu.fi\\[+6pt]
    }
}

\begin{document}

\maketitle
\thispagestyle{empty}
\pagestyle{empty}


\begin{abstract}

    Ultra-wideband (UWB) technology is a mature technology that contested other wireless technologies in the advent of the IoT but did not achieve the same levels of widespread adoption. In recent years, however, with its potential as a wireless ranging and localization solution, it has regained momentum. Within the robotics field, UWB positioning systems are being increasingly adopted for localizing autonomous ground or aerial robots. In the Industrial IoT (IIoT) domain, its potential for ad-hoc networking and simultaneous positioning is also being explored. This survey overviews the state-of-the-art in UWB networking and localization for robotic and autonomous systems. We also cover novel techniques focusing on more scalable systems, collaborative approaches to localization, ad-hoc networking, and solutions involving machine learning to improve accuracy. This is, to the best of our knowledge, the first survey to put together the robotics and IIoT perspectives and to emphasize novel ranging and positioning modalities. We complete the survey with a discussion on current trends and open research problems.

\end{abstract}

\IEEEpeerreviewmaketitle

\section{Introduction}\label{sec:introduction}

Location data is essential in multiple types of autonomous systems, whether it is from an operational perspective (i.e., to enable mobility) or from the point of view of data gathering (i.e., for aggregating data from multiple and/or mobile sources). Specifically, within the robotics domain, accurate localization methods are instrumental for autonomous robots operating in GNSS-denied environments~\cite{rosen4advances}. At the same time, with the rapid development and adoption of wireless technologies in the Industrial IoT (IIoT), high-precision location services are in increasing need~\cite{che2020machine}. Ultimately, simultaneous high-bandwidth wireless data transmission and localization (situated communication) can accelerate the adoption and ubiquity of distributed autonomous systems~\cite{stoy2001using, mohammadmoradi2019srac}.

\begin{figure}
    \centering
    \includegraphics[width=0.49\textwidth]{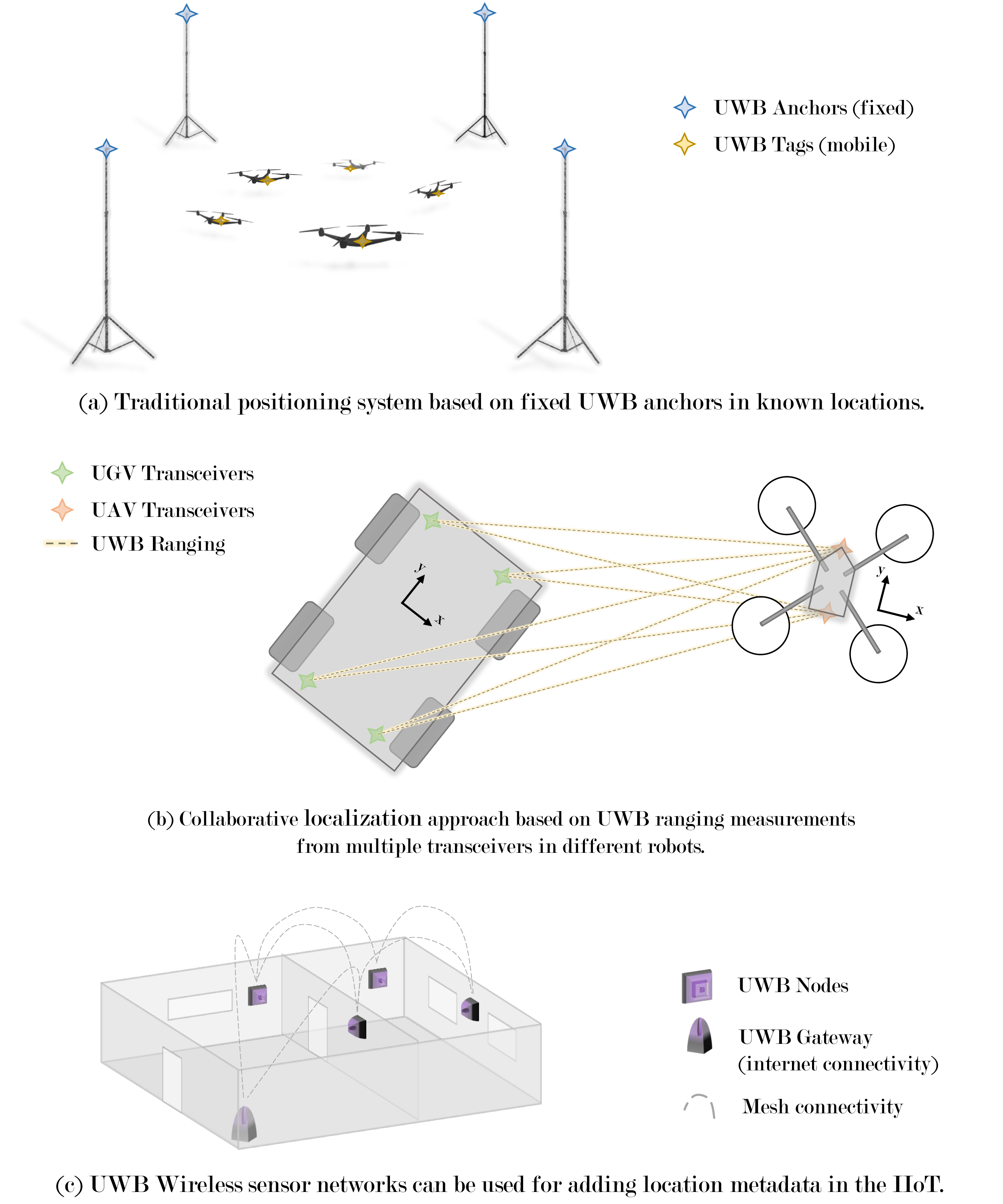}
    \caption{Illustration of different use cases for UWB systems.}
    \vspace{-1em}
    \label{fig:concepual_illustration}
\end{figure}

Wireless ranging technologies have significant benefits as a localization system solution owing to their higher degree of independence to environmental conditions when compared to, e.g., visual sensors~\cite{shule2020uwb}. Among these, ultra-wideband (UWB) wireless technology can provide significantly higher ranging accuracy than Wi-Fi or Bluetooth, among other active radio solutions~\cite{zou2011impulse, ridolfi2018experimental}. Multipath resilience and time resolution are additional advantages~\cite{alarifi2016ultra}. Moreover, state-of-the-art UWB-based positioning systems can achieve, out-of-the-box, localization accuracy and stability levels sufficient for the operation of complex autonomous robots such as micro-aerial vehicles~\cite{queralta2020uwb}, at a fraction of the price of more traditional solutions such as motion capture (MOCAP) systems.

Figure~\ref{fig:concepual_illustration} illustrates three example use-cases of UWB wireless technology. Positioning systems based on a series of fixed nodes in known locations (or anchors), and ranging measurements between these and mobile nodes (or tags), can be used for consistent, long-term localization of mobile robots~\cite{macoir2019uwb}. Alternatively, applications involving multi-robot systems can leverage the use of UWB transceivers onboard different robots for full relative pose estimation~\cite{nguyen2018robust}, and to the level of swarm-wide state estimation even with a single transceiver on each robot~\cite{xu2020decentralized}. The third example then shows a UWB mesh network enabling simultaneously a networking and positioning solution in wireless sensor networks~\cite{mohammadmoradi2019srac}.

Recent surveys and reviews on UWB technology have focused on either the signal processing aspects~\cite{kocur2019signal}, the design of UWB antennas~\cite{soni2020detail, tiwari2018survey}, the potential communication modalities~\cite{fang2016survey}, or the comparison of UWB localization systems with other technologies~\cite{alarifi2016ultra}. Within the robotics field, recent reviews have focused on the different positioning modalities with little attention paid to the technology itself, or the networking possibilities~\cite{shule2020uwb}.

In summary, and to the best of our knowledge, this is the first survey on UWB localization systems and networks that brings together the perspectives of autonomous robotic systems and automation in the IIoT. Moreover, most commercial systems feature mainly fixed anchor-based localization. Nonetheless, in recent years novel solutions have emerged including collaborative localization and more scalable methods. We see a lack of coverage of these approaches in exiting surveys and reviews.

The rest of this survey is organized as follows. Section II introduces novel techniques and recent trends in localization, ad-hoc networking, and scalable UWB systems. Then, Section III presents the state-of-the-art localization systems for autonomous mobile robots. Special attention is paid to aerial robots and multi-robot systems. In Section IV, the review shifts towards the IIoT domain. We discuss the main trends and open problems in Section V, with Section VI concluding the work.

\section{Novel UWB Systems and Approaches}

Different ranging modalities are possible with UWB transceivers, common to the majority of wireless ranging systems. We only briefly describe those that are more relevant within the context of this survey, and refer the reader to previous works for more detailed explanations~\cite{shule2020uwb, chugunov2020toa, zhu2020spin}. Generally, UWB systems based on fixed anchors rely on either time difference of arrival (TDoA) when the anchors are synchronized, or time of flight (ToF) alternatively. TDoA requires a single transmission from a mobile tag to all anchors, whereas ToF measurements are often one-to-one (two way ranging - TWR) and relying on either two or three transmissions (single sided TWR and double sided TWR, respectively). An immediate repercussion of this is that ToF systems are less scalable and require more power, but do not have synchronization requirements. It is also worth noticing that in ToF ranging the positioning is reduced to estimating the intersection of a number of circumferences, while in TDoA systems the localization depends on the intersection of hyperbolas. The former can therefore be more easily extended outside the convex hull defined by the anchor positions, whereas the latter provides more accuracy within the hull but is not extendable outside of it.

\begin{figure}
    \centering
    \includegraphics[width=0.49\textwidth]{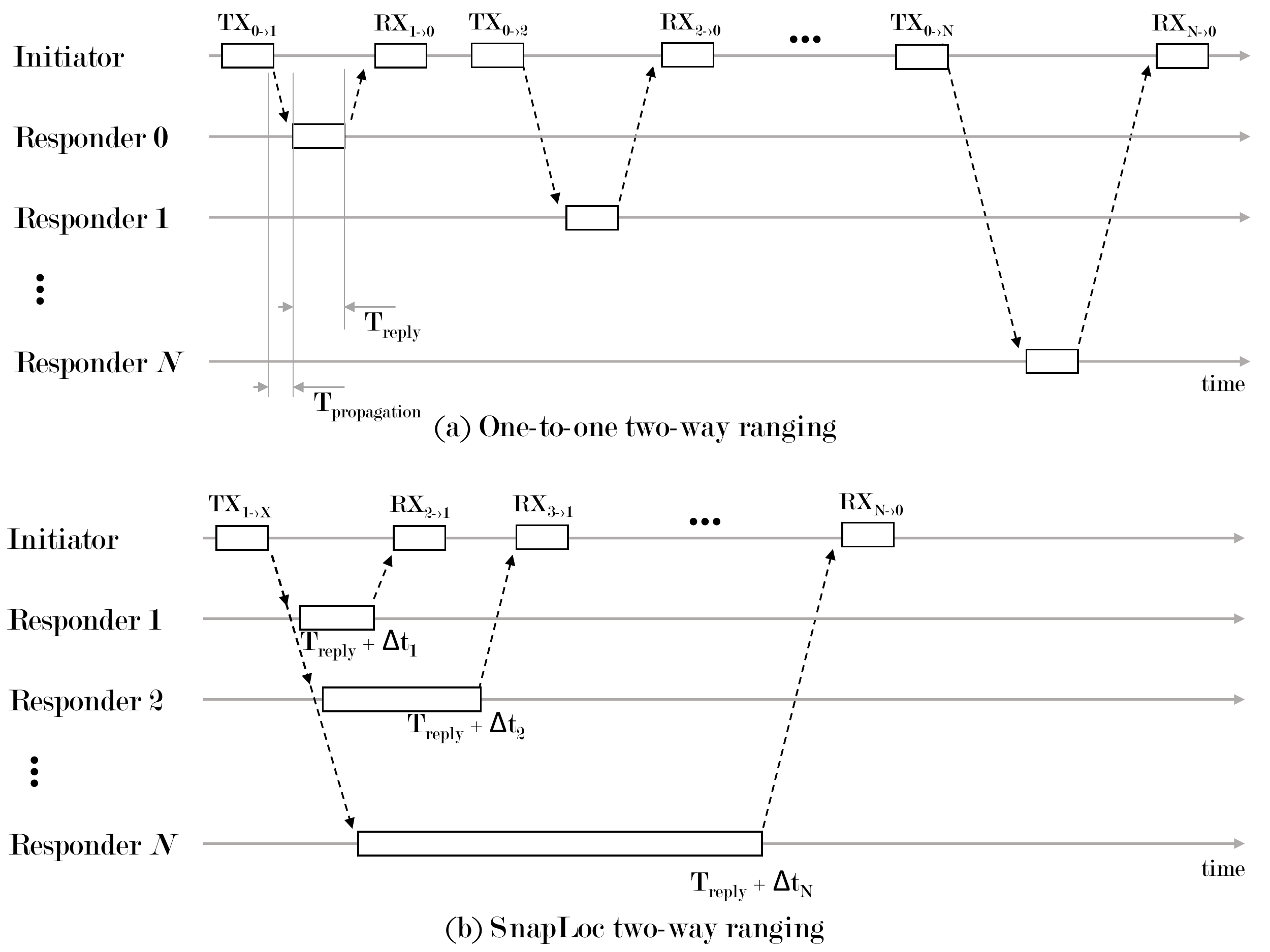}
    \caption{Illustration of two ranging possibilities without node synchronization: (a) typical one-to-one single-sided TWR, and (b) the SnapLoc~\cite{grobetawindhager2019snaploc} reply schedule with predefined delays. The latter system is inherently able to accommodate a significantly larger amount of nodes.}
    \label{fig:snaploc}
\end{figure}

Benefits of UWB-based positioning systems include, e.g., deployment flexibility~\cite{almansa2020autocalibration}, and competitive performance-cost ratio~\cite{queralta2020uwb}. Among the key impediments to wider adoption of UWB systems are, e.g., limited range~\cite{loco} and support for a limited number of nodes in a certain area~\cite{stocker2020towards}. In the following, we review current trends towards more scalable, secure, and extendable systems. We also look into recent literature in ad-hoc UWB networks, and into systems leveraging machine learning techniques for increased positioning accuracy.

\subsection{Scalability}

As real-time localization systems (RTLS) become increasingly ubiquitous, scalability is a key parameter to consider when deploying across large areas. Widely adopted commercial systems, such as Decawave's DRTLS, can support a maximum system-level positioning frequency in the order of the hundreds of herzs. This means that hundreds of tags can be located at 1\,Hz, but only tens of tags at 10\,Hz, thus limiting the real-time nature of the positioning system~\cite{decawave, sewio}.

The research community has therefore dedicated multiple efforts to increase the scalability of UWB systems, from the perspective of high node density~\cite{ridolfi2018analysis}, or faster and concurrent ranging~\cite{grobetawindhager2019snaploc, corbalan2018concurrent, grosswindhager2018concurrent, vecchia2019playing, heydariaan2020anguloc}. Figure~\ref{fig:snaploc} illustrates this concept. Other systems rely on passive tags that use transmissions between anchors and a TDoA scheme to locate themselves~\cite{loco, chen2016network, vecchia2019talla}.


\subsection{Security}

As UWB-based wireless access is inherently leveraged for location purposes, new attack vectors arise beyond cybersecurity threats common to other wireless technologies. These include, for instance, advanced or delayed message replications from malicious nodes which, though not affecting the data itself, mislead the receiver with altered timing, rendering the ranging calculations inaccurate~\cite{stocker2020towards}. In this direction, defining a more secure UWB physical layer is the objective within the IEEE 802.15 WPAN Task Group 4z Enhanced Impulse Radio~\cite{ieee80215z}.

\subsection{Collaborative Localization}

With the majority of commercial UWB positioning systems being based on fixed anchors in known locations, the potential of aggregating single range measurements across distributed systems has gained traction within the robotics field. Particularly suited to multi-robot systems and decentralized and distributed systems, collaborative localization or relative localization approaches have multiple applications.

Relative localization between two robots equipped with multiple transceivers~\cite{nguyen2018robust} can be leveraged for, e.g., autonomous docking~\cite{nguyen2019integrated}, or for collaborative scene reconstruction~\cite{queralta2020vio}. When large numbers of robots are deployed in swarms, UWB ranging and other sensor data can be leveraged for decentralized system-level estate estimation~\cite{xu2020decentralized, qi2020cooperative}. Utilizing time window optimization techniques, these systems can provide accurate relative positioning across the swarm~\cite{xu2021omni}. These approaches can also be extended to track, e.g., people~\cite{zhu2020spin, liu2017cooperative}. Collaborative UWB-based localization in swarms of robots can be leveraged to create experimental platforms that bridge the gap between theory and the real world~\cite{queralta2021towards}. In a different direction, a hybrid system proposed in~\cite{almansa2020autocalibration} relies on anchors for calibrating tags, while the anchors themselves are in turn mobile and collaboratively re-calibrating their positions.

\subsection{Ad-hoc, Mesh and Situated Communication}

The unparalleled positioning capabilities of UWB are the main reason behind its increasing penetration, yet its potential for serving \textit{simultaneously} as a communication and networking means remains mostly untapped. Indeed, most systems in the literature rely on well-established wireless technologies other than UWB for data exchanges~\cite{xu2020decentralized, queralta2020vio, nguyen2020viral}. Nonetheless, integrating basic signaling within UWB ranging transmissions can be leveraged for ad-hoc networking. The integration of UWB ranging for collaborative localization and Wi-Fi mesh networks has also been identified as a potential tool for ad-hoc networking and situated communication in robot swarms~\cite{queralta2021towards}.

In a recent study~\cite{mohammadmoradi2019srac}, a novel approach to simultaneous ranging and communication in UWB networks was presented. Rather than utilizing specific transmissions for ranging, separating them from the data exchange, the authors argue that in most cases ranging can be integrated within nominal network traffic if an immediate calculation is not needed. 

\subsection{Sensor Fusion and ML/DL for Enhanced Accuracy}

Different efforts are being put to increase the accuracy of UWB-based ranging. From a higher-level perspective, with integration within multi-modal sensor fusion systems, and at a lower level with analysis of the received signal to differentiate, e.g., between a line of sight (LOS) and non-LOS (NLOS) transmissions. In the latter category, different machine learning algorithms can be applied to classify between LOS and NLOS transmissions. For instance, a Naive Bayesian algorithm proposed in~\cite{che2020machine} can maintain consistent ranging accuracy in both LOS and NLOS measurements. In~\cite{jiang2020uwb}, a CNN-LSTM deep neural network classifies between LOS and NLOS transmissions. In~\cite{yu2018novel}, a least-squares optimizer is introduced to reduce the NLOS ranging errors.

\section{UWB in Robotics}

Accurate localization in the robotics field is still an open problem across multiple deployment scenarios. Especially in GNSS-denied environments, owing to the different accuracy, availability, area coverage, scalability, cost, and privacy requirements~\cite{Alarifi2016uwbIndoor}. UWB systems are flexible and can operate in mixed indoor-outdoor environments, enabling centimeter-level accuracy localization with either global or relative positioning~\cite{almansa2020autocalibration, prorok2012TDOA}. In this section, we focus on describing how UWB technology has been leveraged in different types of robotic systems and in various application domains. 

\subsection{UWB in Mobile Robots} %

Autonomous mobile robots are already a reality across multiple industries and domains, from home cleaning to in-warehouse transportation and including rapidly emerging autonomous last-mile delivery solutions. In controlled environments where anchors can be installed, UWB systems can benefit autonomous operation aiding in long-term autonomy, persistent localization, and pose initialization, among others. Moreover, it enhances well-established onboard odometry approaches (e.g., lidar- or visual-based) for simultaneous localization and mapping (SLAM), with the external UWB positioning system reducing the drift and accumulated error~\cite{tiemann2018enhanced, song2019uwb}. Consider e.g. extreme cases and feature-less environments such as tunnel-like environments, where lidar-based system will suffer from degenerate geometry. Or similarly, vision sensors in low-light or over-exposure conditions. In these scenarios, even ranging to a single fixed UWB transceiver without full external position estimation can provide significant improvements to the overall localization~\cite{zhen2019estimating}. It is also worth mentioning that UWB positioning systems do not provide orientation when a single tag is installed on a mobile robot. With multiple tags, the mean UWB ranging error can be of the same order of magnitude as the size of the robot, thus excluding the possibility of directly calculating orientation from the position of several tags. In this direction, multiple works in the literature been devoted to investigating different methods and their accuracy to estimating orientation with multiple tags attached on single object~\cite{bonsignori2020estimation, chen2020uwb}. This has also been a topic of study in cooperative relative localization systems involving two or more robots~\cite{theussl2019measurement, s19204366}. For example, a full 6-DOF pose estimator is presented in~\cite{li2020uwb}, where UWB ranging data is fused with with additional sensors (e.g., IMU). 

\subsection{UWB in Aerial Robotics}

Unprecedented advantages in the field of aerial robotics have occurred over the past decade. The first part of it was largely due to the possibilities of MOCAP systems, with advances in visual-inertial-odometry (VIO) and lidar-based SLAM further pushing the frontiers of the field~\cite{chung2018survey}. UWB provides a competitive alternative to MOCAP systems that is accurate and stable enough by itself whenever centimeter-level accuracy suffices~\cite{queralta2020uwb, guo2016ultra, qingqing2021adaptive}. In any case, UWB ranging or localization is rarely used as a standalone system and is very often integrated with IMU data (and/or other sources of odometry or location data) to add orientation estimation and improve the oveall accuracy~\cite{yao2017integrated}. From the perspective of aerial systems, part of the research to date has focused on investigating the properties of ground-to-air UWB-based ranging~\cite{kong2015ground, khawaja2019uwb}. This is an important aspect to consider, as most transceivers include antennas with radiation patterns designed for largely two-dimensional spaces. 

%
Much attention has also been put to integrating UWB within swarms of aerial robots~\cite{xu2020decentralized, qi2020cooperative, li2018swams}. In aerial swarms, external UWB positioning systems supporting multiple nodes are applicable. They can be considered as a competitive replacement, in some scenarios, to GNSS-RTK systems, with simpler deployment and initialization. In any case, most research has shifted to relative estate estimation within the swarm. Compared, e.g., to vision-based relative localization estimation, UWB systems are less dependent on environmental conditions and provide omnidirectional mutual detection.

\subsection{Sensor Fusion in UWB-Based Systems}

While UWB ranging enables accurate positioning, multipath interference, NLOS transmission degrade the positioning performance. Therefore, UWB ranging data is often integrated and fused with different sensors and sources of location, inertial or odometry data. These include raw IMU data~\cite{yao2017integrated}, VIO estimators~\cite{nguyen2019integrated}, or lidar~\cite{song2019uwb}. With single UWB transceivers only providing positions in 2D or 3D spaces, full pose estimation requires fusion with at least IMU. Multiple works have presented different approaches, e.g., with Extended Kalman Filters (EKF), to complete the full pose state estimation~\cite{liu2017cooperative, li2019imuEKF, feng2020kalman}. 
In outdoor environments, GNSS technology is at the core of positioning methods, with unparalleled coverage and effortless integration. However, GNSS accuracy is easily influenced by multipath interference in clustered environments and weak penetration. To achieve seamless positioning in hybrid outdoors-indoors environments, a GPS/UWB/MARG collaborative positioning system is introduced in~\cite{zhang2019combined}.

In general terms, the most significant trend at the moment is in integrating UWB with visual sensors and, in particular, VIO estimators~\cite{nguyen2020viral, xu2020decentralized, xu2021omni, shi2018visual}. In multi-robot systems involving aerial robots, vision-based sensors have typically been used for achieving relative localization and autonomous landing. In~\cite{nguyen2019integrated}, a UWB-Vision fusion method is presented for autonomous docking of UAVs on mobile platforms. The authors designed a relative localization scheme by using distance measurements from sequential ranging to UWB anchors and relative displacement measurements from visual odometers. These two data streams were fed to a recursive least square algorithm to estimate the relative position to the moving target. The final docking was then done based mostly on fiducial markers using vision. From extensive indoor and outdoor experiments show the integrated UWB vision approach can be achieved from a distance up to 50m.

\subsection{UWB in Swarm and Multi-Robot Systems}

Collaborative multi-robot systems have been a recurrent research topics over the past decades~\cite{queralta2020sarmrs}. Applications range from surveillance~\cite{nigam2011control} to dense scene construction~\cite{queralta2020vio}, and including search and rescue~\cite{queralta2020autosos}, among many others. Accurate and robust relative localization is one essential aspect in tasks such as cooperative manipulation, collaborative sensing, exploration, and transportation applications. 
UWB-based relative state estimation method in multi-robot systems can be divided into centralized~\cite{nguyen2018robust, li2018swams} and decentralized~\cite{xu2020decentralized, xu2021omni, queralta2020reconfigurable} approaches. In UWB-based decentralized relative state estimation systems, each robot performs relative state estimation individually and then exchanges such information with other robots.  
Compared with a centralized system where all the relative position estimation is processed on one device or based on an external system, decentralized systems have advantages including~\cite{xu2020decentralized, xu2021omni}: (i) flexibility, i.e., only onboard sensors and computing power are utilized, so no external off-board sensors or computing power is required; and (ii) robustness, i.e., the system can handle the lack of detection results as well as malfunctions of individual robots.
  
Another important consideration that is worth addressing here is signal interference in multi-robot UWB-based positioning systems. In the presence of multiple robots, signals from UWB tags can suffer from multipath transmission being reflected in other robots, in turn causing interference at the receiver. Multiple research efforts have delved into investigating the impact of MUI (Multi-User Interference)~\cite{kristem2014experimental} and mitigating MUI using a coherent receiver in~\cite{kristem2014coherent}.

\subsection{UWB in Human-Robot Interaction}

Within another important research area in robotics, UWB systems have also been employed for human-robot interaction. With the objective of accurately and precisely capturing and recognizing human motion, different systems have been focusing, e.g., in in hand gestures and body movement~\cite{corrales2008hybrid}, or processing electromyography (EMG) signals~\cite{bi2019review}. In a human motion capture system, as has been previously introduced for robots, global positioning from fixed UWB anchors can also be ported to mobile tags attached to the human body. Other sensors such as IMU can then aggregate acceleration and orientation data~\cite{corrales2008hybrid, neto2019gesture}.

\section{UWB in the IIoT}


Owing to features such as high-bandwidth transmission, low power consumption, robustness against interference, and centimeter-level accuracy performance, UWB wireless technology can be applied to multiple aspects of the IoT. Its potential applications include, e.g. in the smart city domain, intelligent transportation systems, or smart buildings~\cite{minoli2018ultrawideband}. From an industrial perspective and within the IIoT, other potential applications include inventory management~\cite{macoir2019uwb}, smart grids~\cite{cheng2020real, khan2014smart}, fleet management~\cite{akhavian2013knowledge}, safety in human-robot operations~\cite{ivvsic2020uwb}, or asset tracking~\cite{li2016tracking}. 

This section focuses on UWB connectivity and positioning in the IIoT, showcasing the literature that demonstrates UWB is a reliable solution for industrial environments. We then introduce a variety of use cases where UWB technology has been applied.


\subsection{Reliable UWB Communication and Positioning in the IIoT}


%
%
%
%
%

UWB networking solutions were first proposed over two decades ago and contested other wireless technologies at the advent of the IoT, after which it lost part of the popularity it is now regaining. Early works analyzing the performance of UWB communication in industrial environments include~\cite{karedal2004statistical}, where the authors present a statistical model to describe the behavior of UWB channels in a factory hall paving the way for a realistic UWB system design. Another early work providing a comprehensive overview of UWB networking and communication systems is available in~\cite{di2006uwb}, which covers the major topics in UWB networking including signal processing, channel measurement, modeling and regulation, and the at the time untapped potential of the technology for positioning. In essence, UWB technology was already relatively mature over a decade ago and crystallized in multiple standards including the IEEE 802.15.4a standard in~\cite{karapistoli2010overview}.

More recently, an experimental study on UWB connectivity in industrial environments~\cite{schmidt2018experimental} concludes that UWB is a promising technology that has lower end-to-end latency, high-bandwidth wireless connectivity, and lower package loss rate compared to to Zigbee. It thus offers competitive advantages and is able to support emerging industrial application, especially in scenarios where a wireless system is a necessity. Within the field of wireless sensor networks, UWB communication has also been considered in different works. For example, in~\cite{schmidt2019study}, the authors introduce the potential of building a self-powered industrial sensor network arguing that UWB can act as an enabler to significantly extend the lifetime of wireless sensor networks. In the area of smart grids, UWB connectivity has been integrated into smart meters~\cite{khan2014smart}, where it provides a more energy-efficient and reliable solution than other wireless technologies that also expose the physical layer for low-level integration.

%
%

UWB connectivity has also been considered within the transportation industry for extreme environments. For instance, in~\cite{ratiu2018wireless}, a UWB-based solution to transmit sensor data wirelessly within a spacecraft has been proposed. A similar approach to transmitting sensor data within aircraft is available in~\cite{neuhold2016poster, neuhold2016experiments}.  In the railway industry, UWB has been proposed as a solution for inter-car connectivity~\cite{fraga2017towards}, safety assurance, and signaling, for example, the surveillance of some critical parts such as level crossing control, interlocking, and dispatching~\cite{govoni2015ultra}. 

In terms of UWB-based localization, multiple examples in the literature demonstrate that UWB positioning systems are more robust and accurate than their Wi-Fi or Bluetooth counterparts which are often applied in industrial environment~\cite{woo2011application, karaagac2017evaluation}. Specifically, UWB positioning has been proven significantly more accurate in harsh and complex environments. The real-time performance of UWB positioning in industrial or manufacturing scenarios of both a low clutter density and a high clutter density has been evaluated and centimeter-level accuracy validated in~\cite{schroeer2018real, razzaghpour2019short}. Industrial environments are widely known for their high degrees of operational complexity, and the performance of UWB in both static and dynamic environments has been studied in~\cite{delamare2020static}. The paper concludes that competitive accuracy levels can be maintained as environmental complexity increases simply by increasing as well the numbers of anchors deployed in the same environment. UWB systems have also been incorporated within hybrid real-time location systems integrating other wireless position technologies to provide more adaptive and robust positioning services in harsh or more intelligent industrial environments.

Within the different UWB positioning solutions, an extensive comparison between common commercial systems is available in~\cite{ruiz2017comparing} where the indoor performance of three different commercially available UWB positioning systems including Ubisense~\cite{ubisense}, BeSpoon~\cite{bespoon} and DecaWave~\cite{decawave_web} has been analyzed.

\subsection{Industrial Use Cases}


Commercial UWB systems ready to be deployed in industrial environments out-of-the-box are now widely available. Popular systems include Sewio~\cite{sewio} or Infsoft~\cite{infsoft}, which can be flexibly deployed in warehouses, production floors, or smart manufacturing workshops. In industrial warehouses, UWB technology has been extensively evaluated~\cite{wang2018research} and its applications to positioning, data transmission~\cite{lee2017iot}  and safety assurance of human-robot collaborative work~\cite{ivvsic2020uwb} have been proposed. In~\cite{macoir2019uwb}, the authors proposed to use UWB localization for drone-based inventory management.



In more remote environments, mine robots have become increasingly instrumental in the mining industry. Owing to the complexity of underground environment, available localization methods are limited. A potential solution introduced in~\cite{li2020uwb} exploits UWB ranging measurements and their fusion with IMU data to generate pseudo-GPS positioning data to a standard controller, with full 6-DOF pose estimation.
In addition, the technology can also be put to use in personnel tracking in underground mine environments~\cite{wu2020research}. In a construction site, UWB technology is a potential solution for increasing productivity, safety monitoring~\cite{teizer2007combined} and construction fleet management~\cite{akhavian2013knowledge}. With regards to precise farming in agricultural environments, UWB technology has potential as a low-cost supplement of GPS for sensor or agricultural robot localization and environment monitoring~\cite{dos2018robot, wu2008localization}.

Finally, it is worth mentioning the applications of UWB technology in pandemic times. Regarding contact tracing, which has been a topic of intense research and development during the COVID-19 pandemic, UWB technology has been considered as a solution for user location tracking, proximity detection, and distance alerts to avoid infection~\cite{shubina2020survey}. In this direction, multiple commercial solutions for ensuring the maintenance of social distancing in industrial environments relying in UWB have emerged~\cite{noccela}. 

%
%

\section{Trends and Open Research Questions}

This section discusses the main research trends in terms of applicability of UWB connectivity and ranging to autonomous systems, and the most important research directions regarding the scalability and security of such systems.

Within the robotics field, there is a growing trend in integrating UWB ranging and localization systems within well-established estate estimation and autonomy stacks~\cite{nguyen2020viral}. In general, fixed UWB systems based on anchor nodes placed in known locations are able to provide a competitive alternative to MOCAP systems or GNSS-RTK systems. While anchor-based systems are indeed becoming increasingly ubiquitous, we see the main research trends being directed towards relative estate estimation in systems comprising multiple robots. Finally, there is also significant traction in the area of exploiting UWB connectivity and not just its ranging capabilities, enabling situated communication in distributed autonomous systems.

From the perspective of scalability, there has been significant traction in recent years building towards the design of methods for more scalable ranging. This is increasing the number of UWB transceivers that can be located in a given area in real-time. At a lower level, this is being done by increasing the concurrency or transmissions and exploiting interference~\cite{grobetawindhager2019snaploc, heydariaan2020anguloc}. At an application level, collaborative localization approaches in distributed and multi-robot systems waive the need for fixed anchors. At the same time, by fusing UWB data with other sensor data, they enable higher accuracy and consistent positioning between UWB ranging estimations~\cite{xu2020decentralized, queralta2020vio, xu2021omni, nguyen2020viral}.

In terms of enhancing the security of UWB networks and positioning systems, there is a clear need for systems that are both \textit{secure} and \textit{scalable}~\cite{stocker2020towards}. This is of paramount importance for a technology that is being applied within safety-critical industrial settings~\cite{sewio, infsoft, huang2017real}, or in autonomous robots requiring high-degrees of reliability~\cite{queralta2020uwb}.

Within the IIoT, we see the most relevant research being directed towards the design, development, and deployment of dependable wireless sensor networks in harsh operational environments. These vary from underground mines to wireless networking aboard spacecraft. Furthermore, UWB positioning systems are providing a competitive solution for asset and personnel tracking, and they can be integrated into different types of intelligent industrial systems ranging from autonomous robots in logistic warehouses to complex machinery in production floors.



%
%

\section{Conclusion}

In this review, we have covered recent developments within the field of UWB-based networking and positioning. Our focus has been, from the technology perspective, on describing novel approaches mainly in the areas of (i) ranging, (ii) networking, and (iii) decentralized localization. From an application point of view, we have covered major advances in the fields of robotics, autonomous systems, and the IIoT. In the discussion section, we have identified some of the key current research directions to be (i) scalability, (ii) security, and (iii) integration within multi-modal sensor fusion systems for localization.

UWB wireless technology did not meet the wide adoption expectations of many with the advent of the IoT, but it is on a clear path to becoming ubiquitous and essential across various industries and application domains. It arguably packs one of the lowest power footprints and the most precise location capabilities within the family of unlicensed wireless technologies. Finally, being able to co-exist with all current radio systems in use, increasingly wider adoption is to be expected.


\section*{Acknowledgment}

This research work is supported by the Academy of Finland's AutoSOS project (Grant No. 328755) and RoboMesh project (Grant No. 336061).



\begin{spacing}{0.97}
\bibliographystyle{IEEEtran}
\bibliography{bibliography-complete}
\end{spacing}

\end{document}